\documentclass[11pt,a4paper]{article}
\usepackage[hyperref]{acl2021}
\usepackage{times}
\usepackage{latexsym}

\usepackage{microtype}

\usepackage{balance}
\usepackage{url}
\usepackage{graphicx}
\usepackage{amsmath,amssymb,amsfonts}
\usepackage{algorithmic}
\usepackage{textcomp}
\usepackage{xcolor}
\usepackage{tikz}
\usetikzlibrary{topaths,calc}
\usepackage[ruled]{algorithm2e}
\usepackage{subfigure}
\usepackage{booktabs}
\usepackage{balance}
\usepackage{colortbl}
\usepackage{enumitem}

\newtheorem{myDef}{Definition}

\aclfinalcopy 
\def\aclpaperid{505} 

\title{Measuring Fine-Grained Domain Relevance of Terms: \\ A Hierarchical Core-Fringe Approach}

\author{Jie Huang$^{1,3}$ $\quad$ Kevin Chen-Chuan Chang$^{1,3}$ $\quad$ Jinjun Xiong$^{2,3}$ $\quad$ Wen-mei Hwu$^{1,3}$ \\
 $^1$University of Illinois at Urbana-Champaign, USA \\
 $^2$IBM Thomas J. Watson Research Center, USA \\
 $^3$IBM-Illinois Center for Cognitive Computing Systems Research (C3SR), USA \\
 \texttt{\{jeffhj, kcchang, w-hwu\}@illinois.edu} \\
 \texttt{jinjun@us.ibm.com}
}

\date{}

\begin{document}
\maketitle
\begin{abstract}
We propose to measure \emph{fine-grained domain relevance}-- the degree that a term is relevant to a broad (e.g., computer science) or narrow (e.g., deep learning) domain. Such measurement is crucial for many downstream tasks in natural language processing. To handle \emph{long-tail terms}, we build a \emph{core-anchored semantic graph}, which uses \emph{core terms} with rich description information to bridge the vast remaining \emph{fringe terms} semantically. To support a \emph{fine-grained domain} without relying on a matching corpus for supervision, we develop \emph{hierarchical core-fringe learning}, which learns core and fringe terms jointly in a semi-supervised manner contextualized in the hierarchy of the domain. To reduce \emph{expensive human efforts}, we employ \emph{automatic annotation} and \emph{hierarchical positive-unlabeled learning}. Our approach applies to big or small domains, covers head or tail terms, and requires little human effort. Extensive experiments demonstrate that our methods outperform strong baselines and even surpass professional human performance.\footnote{The code and data, along with several term lists with domain relevance scores produced by our methods are available at \url{https://github.com/jeffhj/domain-relevance}.}
\end{abstract}

\section{Introduction}

With countless terms in human languages, no one can know all terms, especially those belonging to a technical domain. 
Even for domain experts, it is quite challenging to identify all terms in the domains they are specialized in. 
However, recognizing and understanding domain-relevant terms is the basis to master domain knowledge.
And having a sense of domains that terms are relevant to is an initial and crucial step for term understanding.

In this paper, as our {\bf problem}, we propose to measure \emph{fine-grained domain relevance},
which is defined as the degree that a term is relevant to a given domain, and the given domain can be broad or narrow--
an important property of terms that has not been carefully studied before. 
E.g., \textit{deep learning} is a term relevant to the domains of computer science and, more specifically, machine learning, but not so much to others like database or compiler. Thus, it has a high domain relevance for the former domains but a low one for the latter.
From another perspective, we propose to decouple extraction and evaluation in automatic term extraction that aims to extract domain-specific terms from texts \cite{amjadian2018distributed,hatty2020predicting}. 
This decoupling setting is novel and useful because it is not limited to broad domains where a domain-specific corpus is available, and also does not require terms must appear in the corpus.

A good command of domain relevance of terms will facilitate many downstream applications. 
E.g., to build a domain taxonomy or ontology, a crucial step is to acquire relevant terms \cite{al2019automatic,shang2020taxonomy}.
Also, it can provide or filter necessary candidate terms for domain-focused natural language tasks \cite{huang2020exploring}. 
In addition, for text classification and recommendation, the domain relevance of a document can be measured by that of its terms.

We aim to measure fine-grained domain relevance as a semantic property of any term in human languages.
Therefore, to be practical, the proposed model for domain relevance measuring must meet the following \textbf{requirements}:
1) covering almost all terms in human languages;
2) applying to a wide range of broad and narrow domains; and
3) relying on little or no human annotation.

However, among countless terms, 
only some of them are popular ones organized and associated with rich information on the Web, e.g., Wikipedia pages, which we can leverage to characterize the domain relevance of such ``head terms.''
In contrast, there are numerous ``long-tail terms''-- those not as frequently used-- which lack descriptive information. 
As \textbf{Challenge 1}, how to measure the domain relevance for such \emph{long-tail terms}?

On the other hand, among possible domains of interest, only those broad ones (e.g., physics, computer science) naturally have domain-specific corpora.
Many existing works \cite{velardi2001identification,amjadian2018distributed,hatty2020predicting} have relied on such domain-specific corpora to identify domain-specific terms by contrasting their distributions to general ones. 
In contrast, those fine-grained domains (e.g., quantum mechanics, deep learning)-- which can be any topics of interest-- do not usually have a matching corpus.
As \textbf{Challenge 2}, how to achieve good performance for a \emph{fine-grained domain} without assuming a domain-specific corpus?

Finally, automatic learning usually requires large amounts of training data.  
Since there are countless terms and plentiful domains, human annotation is very time-consuming and laborious. 
As \textbf{Challenge 3}, how to reduce \emph{expensive human efforts} when applying machine learning methods to our problem?

As our solutions, we propose a hierarchical core-fringe domain relevance learning approach that addresses these challenges.
\textbf{First}, to deal with long-tail terms, we design the \textbf{\emph{core-anchored semantic graph}}, which includes \emph{core terms} which have rich description and \emph{fringe terms} without that information. Based on this graph, we can bridge the domain relevance through term relevance and include any term in evaluation. 
\textbf{Second}, to leverage the graph and support fine-grained domains without relying on domain-specific corpora, we propose \textbf{\emph{hierarchical core-fringe learning}}, which learns the domain relevance of core and fringe terms jointly in a semi-supervised manner contextualized in the hierarchy of the domain.
\textbf{Third}, to reduce human effort, we employ \textbf{\emph{automatic annotation}} and \textbf{\emph{hierarchical positive-unlabeled learning}}, which allow to train our model with little even no human effort.

Overall, our framework consists of two processes: 
1) the \emph{offline construction process}, where a domain relevance measuring model is trained by taking a large set of seed terms and their features as input; 
2) the \emph{online query process}, where the trained model can return the domain relevance of query terms by including them in the core-anchored semantic graph.
Our approach applies to a wide range of domains and can handle any query, while nearly no human effort is required.
To validate the effectiveness of our proposed methods, we conduct extensive experiments on various domains with different settings. 
Results show our methods significantly outperform well-designed baselines and even surpass human performance by professionals.

\section{Related Work}

The problem of domain relevance of terms is related to automatic term extraction, which aims to extract domain-specific terms from texts automatically.
Compared to our task, automatic term extraction, where extraction and evaluation are combined, possesses a limited application and has a relatively large dependence on corpora and human annotation, so it is limited to several broad domains and may only cover a small number of terms.
Existing approaches for automatic term extraction can be roughly divided into three categories: linguistic, statistical, and machine learning methods. 
Linguistic methods apply human-designed rules to identify technical/legal terms in a target corpus \cite{handler2016bag,ha2017technicality}.
Statistical methods use statistical information, e.g., frequency of terms, to identify terms from a corpus \cite{frantzi2000automatic,nakagawa2002simple,velardi2001identification,drouin2003term,meijer2014semantic}. 
Machine learning methods learn a classifier, e.g., logistic regression classifier, with manually labeled data \cite{conrado2013machine,fedorenko2014automatic,hatty2017evaluating}.
There also exists some work on automatic term extraction with Wikipedia \cite{vivaldi2012using,wu2012extracting}. However, terms studied there are restricted to terms associated with a Wikipedia page.

Recently, inspired by distributed representations of words \cite{mikolov2013efficient}, methods based on deep learning are proposed and achieve state-of-the-art performance.
\citet{amjadian2016local,amjadian2018distributed} design supervised learning methods by taking the concatenation of domain-specific and general word embeddings as input. 
\citet{hatty2020predicting} propose a multi-channel neural network model that leverages domain-specific and general word embeddings.

The techniques behind our hierarchical core-fringe learning methods are related to research on graph neural networks (GNNs) \cite{kipf2017semi,hamilton2017inductive}; hierarchical text classification \cite{vens2008decision,wehrmann2018hierarchical,zhou2020hierarchy}; and positive-unlabeled learning \cite{liu2003building,elkan2008learning,bekker2020learning}.

\begin{figure*}[tp!]
\centerline{\includegraphics[width=0.75\linewidth]{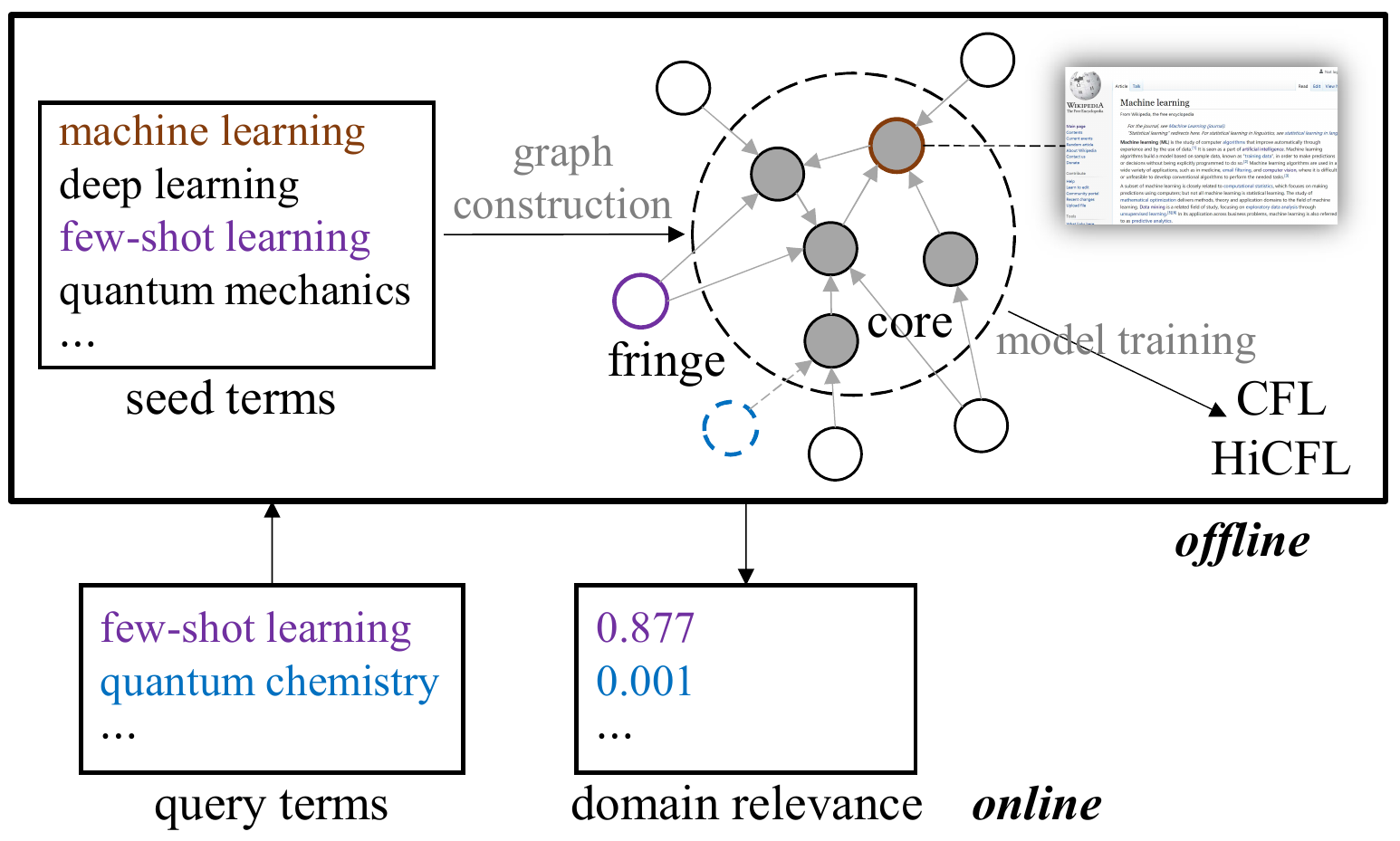}}
\caption{The overview of the framework. 
In this figure, \textit{machine learning} is a core term associated with a Wikipedia page,
\textit{few-shot learning} is a fringe term included in the offline core-anchored semantic graph, and \textit{quantum chemistry} is a fringe term included in the online process. Best viewed in color.}
\label{fig:query}
\end{figure*}

\section{Methodology}

We study the \textit{Fine-Grained Domain Relevance} of terms, which is defined as follows:
\begin{myDef}{(Fine-Grained Domain Relevance)} 
The fine-grained domain relevance of a term is the degree that the term is relevant to a given domain, and the given domain can be broad or narrow.
\end{myDef}

The domain relevance of terms depends on many factors. 
In general, a term with higher semantic relevance, broader meaning scope, and better usage possesses a higher domain relevance regarding the target domain. 
To measure the fine-grained domain relevance of terms, we propose a hierarchical core-fringe approach, which includes an offline training process and can handle any query term in evaluation. The overview of the framework is illustrated in Figure \ref{fig:query}.

\subsection{Core-Anchored Semantic Graph}
\label{sec:graph_construction}

There exist countless terms in human languages; thus it is impractical to include all terms in a system initially. 
To build the offline system, we need to provide seed terms,
which can come from knowledge bases or be extracted from broad, large corpora by existing term/phrase extraction methods \cite{handler2016bag,shang2018automated}. 

In addition to providing seed terms, we should also give some knowledge to machines so that they can differentiate whether a term is domain-relevant or not. 
To this end, we can leverage the description information of terms.
For instance, Wikipedia contains a large number of terms (the surface form of page titles), 
where each term is associated with a Wikipedia article page.  
With this page information, humans can easily judge whether a term is domain-relevant or not. In Section~\ref{sec:automatic_annotation}, we will show the labeling can even be done completely automatically.

However, considering the countless terms, the number of terms that are well-organized and associated with rich description is small. 
How to measure the fine-grained domain relevance of terms without rich information is quite challenging for both machines and humans.

Fortunately, terms are not isolated, while complex relations exist between them. 
If a term is relevant to a domain, it must also be relevant to some domain-relevant terms and vice versa. 
This is to say, we can bridge the domain relevance of terms through term relevance.
Summarizing the observations, we divide terms into two categories: \emph{core terms}, which are terms associated with rich description information, e.g., Wikipedia article pages, and \emph{fringe terms}, which are terms without that information.
We assume, for each term, there exist some relevant core terms that share similar domains. 
If we can find the most relevant core terms for a given term, its domain relevance can be evaluated with the help of those terms. 
To this end, we can utilize the rich information of core terms for ranking.

Taking Wikipedia as an example, each core term is associated with an article page, so they can be returned as the ranking results (result term) for a given term (query term).
Considering the data resources, we use the built-in Elasticsearch based Wikipedia search engine\footnote{\url{https://en.wikipedia.org/w/index.php?search}} \cite{gormley2015elasticsearch}. 
More specifically, we set the maximum number of links as $k$ ($5$ as default). For a query term $v$, i.e., any seed term, we first achieve the top $2k$ Wikipedia pages with exact match. For each result term $u$ in the core, we create a link from $u$ to $v$. If the number of links is smaller than $k$, we do this process again without exact match and build additional links.
Finally, we construct a term graph, named \textbf{Core-Anchored Semantic Graph}, where nodes are terms and edges are links between terms. 

In addition, for terms that are not provided initially, we can also handle them as fringe terms and connect them to core terms in evaluation. 
In this way, we can include any term in the graph.

\subsection{Hierarchical Core-Fringe Learning}
\label{sec:learning}

In this section, we aim to design learning methods to learn the fine-grained domain relevance of core and fringe terms jointly.
In addition to using the term graph, we can achieve features of both core and fringe terms based on their linguistic and statistical properties \cite{terryn2019no,conrado2013machine} or distributed representations \cite{mikolov2013distributed,yu2015learning}. 
We assume the labels, i.e., domain-relevant or not, of core terms are available, which can be achieved by an automatic annotation mechanism introduced in Section~\ref{sec:automatic_annotation}.

As stated above, 
if a term is highly relevant to a given domain, it must also be highly relevant to some other terms with a high domain relevance and vice versa. 
Therefore, to measure the domain relevance of a term, in addition to using its own features, we aggregate its neighbors' features.
Specifically, we propagate the features of terms via the term graph and use the label information of core terms for supervision. 
In this way, core and fringe terms help each other, and the domain relevance is learned jointly.
The propagation process can be achieved by graph convolutions \cite{hammond2011wavelets}.
We first apply the vanilla graph convolutional networks (GCNs) \cite{kipf2017semi} in our framework. 
The graph convolution operation (GCNConv) at the $l$-th layer is formulated as the following aggregation and update process:
\begin{equation}
\label{eq:gcnconv}
\boldsymbol{h}_i^{(l+1)} = \phi \Big( \sum_{j \in \mathcal{N}_i \cup \{i\}}{\frac{1}{c_{ij}} \boldsymbol{W}_c^{(l)} \boldsymbol{h}_j^{(l)} + \boldsymbol{b}_c^{(l)} } \Big),
\end{equation}
where $\mathcal{N}_i$ is the neighbor set of node $i$. 
$c_{ij}$ is the normalization constant.
$\boldsymbol{h}_j^{(l)} \in \mathbb{R}^{d^{(l)} \times 1}$ is the hidden state of node $j$ at the $l$-th layer, with $d^{(l)}$ being the number of units; $\boldsymbol{h}_j^{(0)} = \boldsymbol{x}_j$, which is the feature vector of node $j$. $\boldsymbol{W}_c^{(l)} \in \mathbb{R}^{d^{(l+1)} \times d^{(l)}}$ is the trainable weight matrix at the $l$-th layer, and $\boldsymbol{b}_c^{(l)}$ is the bias vector. 
$\phi(\cdot)$ is the nonlinearity activation function, e.g., $\text{ReLU}(\cdot) = \max(0,\cdot)$.

Since core terms are labeled as domain-relevant or not, we can use the labels to calculate the loss:
\begin{equation}
\label{eq:loss1}
\mathcal{L} = - \sum_{i \in \mathcal{V}_{core}}{ ( y_i \log z_i + (1-y_i) \log (1-z_i) )},
\end{equation}
where $y_i$ is the label of node $i$ regarding the target domain, and $z_i = \sigma(h_i^{o})$, with $h_i^{o}$ being the output of the last GCNConv layer for node $i$ and $\sigma(\cdot)$ being the sigmoid function. The weights of the model are trained by minimizing the loss. 
The relative domain relevance is obtained as $\boldsymbol{s} = \boldsymbol{z}$.

Combining with the overall framework, we get the first domain relevance measuring model, \textbf{CFL}, i.e., \textbf{C}ore-\textbf{F}ringe Domain Relevance \textbf{L}earning. 

CFL is useful to measure the domain relevance for broad domains such as computer science.
For domains with relatively narrow scopes, e.g., machine learning, we can also leverage the label information of domains at the higher level of the hierarchy, e.g., CS $\to$ AI $\to$ ML, which is based on the idea that a domain-relevant term regarding the target domain should also be relevant to the parent domain.
Inspired by related work on hierarchical multi-label classification \cite{vens2008decision,wehrmann2018hierarchical}, we introduce a hierarchical learning method considering both global and local information.

We first apply $l_c$ GCNConv layers according to Eq.~\eqref{eq:gcnconv} and get the output of the last GCNConv layer, which is $\boldsymbol{h}_i^{(l_c)}$. 
In order not to confuse, we omit the subscript that identifies the node number.
For each domain in the hierarchy, we introduce a hierarchical global activation $\boldsymbol{a}_p$. The activation at the $(l+1)$-th level of the hierarchy is given as
\begin{equation}
\boldsymbol{a}_p^{(l+1)} = \phi ( \boldsymbol{W}_p^{(l)} [\boldsymbol{a}_p^{(l)};\boldsymbol{h}^{(l_c)}] + \boldsymbol{b}_p^{(l)} ),
\end{equation}
where $[\cdot;\cdot]$ indicates the concatenation of two vectors; $\boldsymbol{a}_p^{(1)} = \phi ( \boldsymbol{W}_p^{(0)} \boldsymbol{h}^{(l_c)} + \boldsymbol{b}_p^{(0)} )$.
The global information is produced after a fully connected layer:
\begin{equation}
\boldsymbol{z}_p = \sigma ( \boldsymbol{W}_p^{(l_p)} \boldsymbol{a}_p^{(l_p)} + \boldsymbol{b}_p^{(l_p)} ),
\end{equation}
where $l_p$ is the total number of hierarchical levels.

To achieve the local information for each level of the hierarchy, the model first generates the local hidden state $\boldsymbol{a}_q^{(l)}$ by a fully connected layer:
\begin{equation}
\boldsymbol{a}_q^{(l)} = \phi ( \boldsymbol{W}_t^{(l)} \boldsymbol{a}_p^{(l)} + \boldsymbol{b}_t^{(l)}).
\end{equation}
The local information at the $l$-th level of the hierarchy is then produced as
\begin{equation}
\boldsymbol{z}_q^{(l)} = \sigma ( \boldsymbol{W}_q^{(l)} \boldsymbol{a}_q^{(l)} + \boldsymbol{b}_q^{(l)}).
\end{equation}

In our core-fringe framework, all the core terms are labeled at each level of the hierarchy. 
Therefore, the loss of hierarchical learning is computed as
\begin{equation}
\mathcal{L}_h = \epsilon(\boldsymbol{z}_p, \boldsymbol{y}^{(l_p)}) + \sum_{l=1}^{l_p} \epsilon(\boldsymbol{z}_q^{(l)}, \boldsymbol{y}^{(l)}),
\end{equation}
where $\boldsymbol{y}^{(l)}$ denotes the labels regarding the domain at the $l$-th level of the hierarchy and $\epsilon(\boldsymbol{z}, \boldsymbol{y})$ is the binary cross-entropy loss described in Eq.~\eqref{eq:loss1}. 
In testing,
The relative domain relevance $\boldsymbol{s}$ is calculated as
\begin{equation}
\boldsymbol{s} = \alpha \cdot \boldsymbol{z}_p + (1-\alpha) \cdot (\boldsymbol{z}_q^{(1)} \circ \boldsymbol{z}_q^{(2)}, ..., \boldsymbol{z}_q^{(l_p)}),
\end{equation}
where $\circ$ denotes element-wise multiplication. $\alpha$ is a hyperparameter to balance the global and local information ($0.5$ as default). 
Combining with our general framework, we refer to this model as \textbf{HiCFL}, i.e., \textbf{Hi}erarchical \textbf{CFL}.

{\flushleft \textbf{Online Query Process}.}
If seed terms are provided by extracting from broad, large corpora relevant to the target domain, most terms of interest will be already included in the offline process. 
In evaluation, for terms that are not provided initially, our model treats them as fringe terms. Specifically, when receiving such a term, the model connects it to core terms by the method described in Section~\ref{sec:graph_construction}. With its features (e.g., compositional term embeddings) or only its neighbors' features (when features cannot be generated directly), the trained model can return the domain relevance of any query.

\subsection{Automatic Annotation and Hierarchical Positive-Unlabeled Learning}
\label{sec:automatic_annotation}

{\flushleft \textbf{Automatic Annotation}.}
For the fine-grained domain relevance problem, human annotation is very time-consuming and laborious because the number of core terms is very large regarding a wide range of domains.
Fortunately, in addition to building the term graph, we can also leverage the rich information of core terms for automatic annotation.

In the core-anchored semantic graph constructed with Wikipedia, each core term is associated with a Wikipedia page,
and each page is assigned one or more categories. All the categories form a hierarchy, furthermore providing a category tree. 
For a given domain, we can first traverse from a root category and collect some gold subcategories. For instance, for computer science, we treat \textit{category: subfields of computer science}\footnote{\url{https://en.wikipedia.org/wiki/Category:Subfields_of_computer_science}} as the root category and take categories at the first three levels of it as gold subcategories. Then we collect categories for each core term and examine whether the term itself or one of the categories is a gold subcategory. If so, we label the term as positive. 
Otherwise, we label it as negative. 
We can also combine gold subcategories from some existing domain taxonomies and extract the categories of core terms from the text description, which usually contains useful text patterns like ``x \textit{is a subfield of} y''.

{\flushleft \textbf{Hierarchical Positive-Unlabeled Learning}.}
According to the above methods, we can learn the fine-grained domain relevance of terms for any domain as long as we can collect enough gold subcategories for that domain. 
However, for domains at the low level of the hierarchy, e.g., deep learning, a category tree might not be available in Wikipedia. To deal with this issue, we apply our learning methods in a positive-unlabeled (PU) setting \cite{bekker2020learning}, where only a small number of terms, e.g., 10, are labeled as positive, and all the other terms are unlabeled. We use this setting based on the following consideration: if a user is interested in a specific domain, it is quite easy for her to give some important terms relevant to that domain.

Benefiting from our hierarchical core-fringe learning approach, we can still obtain labels for domains at the high level of the hierarchy with the automatic annotation mechanism. Therefore, all the negative examples of the last labeled hierarchy can be used as reliable negatives for the target domain. For instance, if the target domain is deep learning, which is in the CS $\to$ AI $\to$ ML $\to$ DL hierarchy, we consider all the non-ML terms as the reliable negatives for DL. 
Taking the positively labeled examples and the reliable negatives for supervision, we can learn the domain relevance of terms by our proposed HiCFL model contextualized in the hierarchy of the domain.

\section{Experiments}
\label{sec:exp}

In this section, we evaluate our model from different perspectives. 
1) We \textbf{compare with baselines} by treating some labeled terms as queries.
2) We \textbf{compare with human professionals} by letting humans and machines judge which term in a query pair is more relevant to a target domain.
3) We conduct intuitive \textbf{case studies} by ranking terms according to their domain relevance.

\subsection{Experimental Setup}

{\flushleft \textbf{Datasets and Preprocessing}.}
To build the system, for offline processing, we extract seed terms from the arXiv dataset (version 6)\footnote{\url{https://www.kaggle.com/Cornell-University/arxiv}}.
As an example, for computer science or its sub-domains, we collect the abstracts in computer science according to the arXiv Category Taxonomy\footnote{\url{https://arxiv.org/category_taxonomy}}, 
and apply \textit{phrasemachine} to extract terms \cite{handler2016bag} with lemmatization and several filtering rules: $\text{frequency} > 10$; $\text{length} \leq 6$; only contain letters, numbers, and hyphen; not a stopword or a single letter. 

We select three broad domains, including 
\textit{computer science} (CS), \textit{physics} (Phy), and \textit{mathematics} (Math); 
and three narrow sub-domains of them, including \textit{machine learning} (ML), \textit{quantum mechanics} (QM), and \textit{abstract algebra} (AA),
with the hierarchies CS $\to$ AI $\to$ ML, Phy $\to$ mechanics $\to$ QM, and Math $\to$ algebra $\to$ AA.
Each broad domain and its sub-domains share seed terms because they share a corpus.
To achieve gold subcategories for automatic annotation (Section~\ref{sec:automatic_annotation}), we collect subcategories at the first three levels of a root category (e.g., \textit{category: subfields of physics}) for broad domains (e.g., physics); or the first two levels for narrow domains, e.g., \textit{category: machine learning} for machine learning. 
Table~\ref{table:dataset} reports the total sizes and the ratios that are core terms.

{\flushleft \textbf{Baselines}.}
Since our task on fine-grained domain relevance is new, there is no existing baseline for model comparison. We adapt the following models on relevant tasks in our setting with additional inputs (e.g., domain-specific corpora):
\begin{itemize}[noitemsep,nolistsep,leftmargin=*]
	\item \textbf{Relative Domain Frequency (RDF)}:
	Since domain-relevant terms usually occur more in a domain-specific corpus, we apply a statistical method using $\text{freq}_s(w)/\text{freq}_g(w)$ to measure the domain relevance of term $w$, where $\text{freq}_s(\cdot)$ and $\text{freq}_g(\cdot)$ denote the frequency of occurrence in the domain-specific/general corpora respectively.
	\item \textbf{Logistic Regression (LR)}: Logistic regression is a standard supervised learning method. We use core terms with labels (domain-relevant or not) as training data, where features are term embeddings trained by a general corpus.
	\item \textbf{Multilayer Perceptron (MLP)}: MLP is a standard neural neural-based model. 
	We train MLP using embeddings trained with a domain-specific corpus or a general corpus as term features, respectively. 
	We also concatenate the two embeddings as features \cite{amjadian2016local,amjadian2018distributed}.
	\item \textbf{Multi-Channel (MC)}: Multi-Channel \cite{hatty2020predicting} is the state-of-the-art model for automatic term extraction, which is based on a multi-channel neural network that takes domain-specific and general corpora as input.
\end{itemize}

\begin{table}[tp!]
    \begin{center}
    \begin{tabular}{c|c|r|r}
        \toprule
        \multicolumn{2}{c|}{domain} & \#terms & core ratio \\
        \midrule
        CS & ML & 113,038 & 27.7\%  \\
        \hline
        Phy & QM & 416,431 & 12.1\%  \\
        \hline
        Math & AA & 103,984 & 26.4\%  \\
        \bottomrule
    \end{tabular}
    \end{center}
    \caption{The statistics of the data.}
    \label{table:dataset}
\end{table}

\begin{table*}[ht]
\begin{center}
\begin{tabular}{c|c|cc|cc|cc}
\toprule
\multicolumn{2}{c|}{} & \multicolumn{2}{c|}{Computer Science} & \multicolumn{2}{c|}{Physics} & \multicolumn{2}{c}{Mathematics} \\
\multicolumn{2}{c|}{} & ROC-AUC & PR-AUC & ROC-AUC & PR-AUC & ROC-AUC & PR-AUC \\
\midrule 
RDF & SG & 0.714 & 0.417 & 0.736 & 0.496 & 0.694 & 0.579 \\
\hline
LR & G & 0.802\scriptsize{$\pm$0.000} & 0.535\scriptsize{$\pm$0.000} & 0.822\scriptsize{$\pm$0.000} & 0.670\scriptsize{$\pm$0.000} & 0.854\scriptsize{$\pm$0.000} & 0.769\scriptsize{$\pm$0.000} \\ 
MLP & S & 0.819\scriptsize{$\pm$0.003} & 0.594\scriptsize{$\pm$0.003} & 0.853\scriptsize{$\pm$0.001} & 0.739\scriptsize{$\pm$0.004} & 0.868\scriptsize{$\pm$0.000} & 0.803\scriptsize{$\pm$0.001} \\
MLP & G & 0.863\scriptsize{$\pm$0.001} & 0.674\scriptsize{$\pm$0.002} & 0.874\scriptsize{$\pm$0.001} & 0.761\scriptsize{$\pm$0.003} & 0.904\scriptsize{$\pm$0.001} & 0.846\scriptsize{$\pm$0.002} \\
MLP & SG & 0.867\scriptsize{$\pm$0.001} & 0.667\scriptsize{$\pm$0.002} & 0.875\scriptsize{$\pm$0.001} & 0.765\scriptsize{$\pm$0.002} & 0.904\scriptsize{$\pm$0.001} & 0.843\scriptsize{$\pm$0.003} \\ 
MC & SG & 0.868\scriptsize{$\pm$0.002} & 0.664\scriptsize{$\pm$0.006} & 0.877\scriptsize{$\pm$0.003} & 0.768\scriptsize{$\pm$0.004} & 0.903\scriptsize{$\pm$0.001} & 0.843\scriptsize{$\pm$0.002} \\
\hline
\textbf{CFL} & G & \textbf{0.885}\scriptsize{$\pm$0.001} & \textbf{0.712}\scriptsize{$\pm$0.002} & \textbf{0.905}\scriptsize{$\pm$0.000} & \textbf{0.812}\scriptsize{$\pm$0.002} & 0.918\scriptsize{$\pm$0.001} & 0.870\scriptsize{$\pm$0.002} \\
\textbf{CFL} & C & 0.883\scriptsize{$\pm$0.001} & 0.708\scriptsize{$\pm$0.002} & 0.901\scriptsize{$\pm$0.000} & 0.800\scriptsize{$\pm$0.001} & \textbf{0.919}\scriptsize{$\pm$0.001} & \textbf{0.879}\scriptsize{$\pm$0.002} \\
\bottomrule
\multicolumn{8}{l}{S and G indicate the corpus used. S: domain-specific corpus, G: general corpus, SG: both.} \\
\multicolumn{8}{l}{C means the pre-trained compositional GloVe embeddings are used.}
\end{tabular}
\end{center}
\caption{Results for broad domains.}
\label{table:results1}
\end{table*}

{\flushleft \textbf{Training}.}
For all supervised learning methods, we apply automatic annotation in Section~\ref{sec:automatic_annotation}, i.e., we automatically label all the core terms for model training.
In the PU setting, we remove labels on target domains. Only 20 (10 in the case studies) domain-relevant core terms are randomly selected as the positives, with the remaining terms unlabeled.
In training, all the negative examples at the previous level of the hierarchy are used as reliable negatives.

{\flushleft \textbf{Implementation Details}.}
Though our proposed methods are independent of corpora, some baselines (e.g., MC) require term embeddings trained from general/domain-specific corpora. 
For easy and fair comparison, we adopt the following approach to generate term features.
We consider each term as a single token, and apply word2vec CBOW \cite{mikolov2013efficient} with negative sampling, where dimensionality is $100$, window size is $5$, and number of negative samples is $5$. 
The training corpus can be a general one (the entire arXiv corpus, denoted as G), or a domain-specific one (the sub-corpus in the branch of the corresponding domain, denoted as S).
We also apply compositional GloVe embeddings \cite{pennington2014glove} (element-wise addition of the pre-trained 100d word embeddings, denoted as C) as non-corpus-specific features of terms for reference.

For all the neural network-based models, we use Adam \cite{kingma2014adam} with learning rate of $0.01$ for optimization, and adopt a fixed hidden dimensionality of $256$ and a fixed dropout ratio of $0.5$. For the learning part of CFL and HiCFL, we apply two GCNConv layers and use the symmetric graph for training.
To avoid overfitting, we adopt batch normalization \cite{ioffe2015batch} right after each layer (except for the output layer) and before activation and apply dropout \cite{hinton2012improving} after the activation.
We also try to add regularizations for MLP and MC with full-batch or mini-batch training, and select the best architecture. 
To construct the core-anchored semantic graph, we set $k$ as $5$.
All experiments are run on an NVIDIA Quadro RTX 5000 with 16GB of memory under the PyTorch framework. The training of CFL for the CS domain can finish in 1 minute.

We report the mean and standard deviation of the test results corresponding to the best validation results with 5 different random seeds.

\begin{table*}[ht]
\begin{center}
\begin{tabular}{c|c|cc|cc|cc}
\toprule
\multicolumn{2}{c|}{} &\multicolumn{2}{c|}{Machine Learning} & \multicolumn{2}{c|}{Quantum Mechanics} & \multicolumn{2}{c}{Abstract Algebra} \\
\multicolumn{2}{c|}{} & ROC-AUC & PR-AUC & ROC-AUC & PR-AUC & ROC-AUC & PR-AUC \\
\midrule 
LR & G & 0.917\scriptsize{$\pm$0.000} & 0.346\scriptsize{$\pm$0.000} & 0.879\scriptsize{$\pm$0.000} & 0.421\scriptsize{$\pm$0.000} & 0.872\scriptsize{$\pm$0.000} & 0.525\scriptsize{$\pm$0.000} \\ 
MLP & S & 0.902\scriptsize{$\pm$0.001} & 0.453\scriptsize{$\pm$0.009} & 0.903\scriptsize{$\pm$0.001} & 0.545\scriptsize{$\pm$0.004} & 0.910\scriptsize{$\pm$0.000} & 0.641\scriptsize{$\pm$0.007} \\
MLP & G & 0.932\scriptsize{$\pm$0.001} & 0.562\scriptsize{$\pm$0.010} & 0.922\scriptsize{$\pm$0.001} & 0.587\scriptsize{$\pm$0.014} & 0.923\scriptsize{$\pm$0.000} & 0.658\scriptsize{$\pm$0.006} \\
MLP & SG & 0.928\scriptsize{$\pm$0.001} & 0.574\scriptsize{$\pm$0.011} & 0.923\scriptsize{$\pm$0.000} & 0.574\scriptsize{$\pm$0.007} & 0.925\scriptsize{$\pm$0.001} & 0.673\scriptsize{$\pm$0.004} \\ 
MC & SG & 0.928\scriptsize{$\pm$0.002} & 0.554\scriptsize{$\pm$0.007} & 0.924\scriptsize{$\pm$0.001} & 0.590\scriptsize{$\pm$0.003} & 0.924\scriptsize{$\pm$0.001} & 0.685\scriptsize{$\pm$0.005} \\
\hline
\textbf{CFL} & G & 0.950\scriptsize{$\pm$0.002} & 0.627\scriptsize{$\pm$0.013} & 0.950\scriptsize{$\pm$0.000} & 0.678\scriptsize{$\pm$0.003} & 0.938\scriptsize{$\pm$0.001} & 0.751\scriptsize{$\pm$0.009} \\
\textbf{HiCFL} & G & \textbf{0.965}\scriptsize{$\pm$0.003} & \textbf{0.645}\scriptsize{$\pm$0.014} & \textbf{0.957}\scriptsize{$\pm$0.001} & \textbf{0.691}\scriptsize{$\pm$0.003} & \textbf{0.942}\scriptsize{$\pm$0.002} & \textbf{0.769}\scriptsize{$\pm$0.006} \\
\bottomrule
\multicolumn{8}{l}{S and G indicate the corpus used. S: domain-specific corpus, G: general corpus, SG: both.} 
\end{tabular}
\end{center}
\caption{Results for narrow domains.}
\label{table:results2}
\end{table*}

\begin{table*}[ht]
\begin{center}
\begin{tabular}{c|c|cc|cc|cc}
\toprule
\multicolumn{2}{c|}{} &\multicolumn{2}{c|}{Machine Learning} & \multicolumn{2}{c|}{Quantum Mechanics} & \multicolumn{2}{c}{Abstract Algebra} \\
\multicolumn{2}{c|}{} & ROC-AUC & PR-AUC & ROC-AUC & PR-AUC & ROC-AUC & PR-AUC \\
\midrule 
LR & G & 0.860\scriptsize{$\pm$0.000} & 0.206\scriptsize{$\pm$0.000} & 0.788\scriptsize{$\pm$0.000} & 0.280\scriptsize{$\pm$0.000} & 0.833\scriptsize{$\pm$0.000} & 0.429\scriptsize{$\pm$0.000} \\ 
MLP & S & 0.804\scriptsize{$\pm$0.003} & 0.144\scriptsize{$\pm$0.003} & 0.767\scriptsize{$\pm$0.009} & 0.260\scriptsize{$\pm$0.005} & 0.804\scriptsize{$\pm$0.006} & 0.421\scriptsize{$\pm$0.010} \\
MLP & G & 0.836\scriptsize{$\pm$0.005} & 0.234\scriptsize{$\pm$0.016} & 0.813\scriptsize{$\pm$0.006} & 0.295\scriptsize{$\pm$0.011} & 0.842\scriptsize{$\pm$0.003} & 0.467\scriptsize{$\pm$0.011} \\
MLP & SG & 0.844\scriptsize{$\pm$0.003} & 0.230\scriptsize{$\pm$0.015} & 0.796\scriptsize{$\pm$0.008} & 0.291\scriptsize{$\pm$0.011} & 0.839\scriptsize{$\pm$0.006} & 0.463\scriptsize{$\pm$0.013} \\ 
MC & SG & 0.852\scriptsize{$\pm$0.006} & 0.251\scriptsize{$\pm$0.019} & 0.795\scriptsize{$\pm$0.014} & 0.303\scriptsize{$\pm$0.017} & 0.861\scriptsize{$\pm$0.004} & 0.547\scriptsize{$\pm$0.006} \\
\hline
\textbf{CFL} & G & 0.918\scriptsize{$\pm$0.001} & 0.441\scriptsize{$\pm$0.009} & \textbf{0.897}\scriptsize{$\pm$0.002} & 0.408\scriptsize{$\pm$0.004} & 0.887\scriptsize{$\pm$0.002} & 0.563\scriptsize{$\pm$0.018} \\
\textbf{HiCFL} & G & \textbf{0.940}\scriptsize{$\pm$0.008} & \textbf{0.508}\scriptsize{$\pm$0.026} & \textbf{0.897}\scriptsize{$\pm$0.004} & \textbf{0.421}\scriptsize{$\pm$0.014} & \textbf{0.915}\scriptsize{$\pm$0.002} & \textbf{0.648}\scriptsize{$\pm$0.009} \\
\bottomrule
\end{tabular}
\end{center}
\caption{Results for narrow domains (PU learning).}
\label{table:results3}
\end{table*}

\begin{table}[ht]
\begin{center}
\begin{tabular}{c|c|c|c}
\toprule
\multicolumn{2}{c|}{} & PR-AUC & PR-AUC (\textbf{PU}) \\
\midrule 
LR & G & 0.509\scriptsize{$\pm$0.000} & 0.449\scriptsize{$\pm$0.000} \\ 
MLP & S & 0.550\scriptsize{$\pm$0.017} & 0.113\scriptsize{$\pm$0.010} \\
MLP & G & 0.586\scriptsize{$\pm$0.016} & 0.299\scriptsize{$\pm$0.027} \\
MLP & SG & 0.590\scriptsize{$\pm$0.005} & 0.217\scriptsize{$\pm$0.013} \\ 
MC & SG & 0.603\scriptsize{$\pm$0.016} & 0.281\scriptsize{$\pm$0.012} \\
\hline
\textbf{CFL} & G & 0.703\scriptsize{$\pm$0.017} & 0.525\scriptsize{$\pm$0.013} \\
\textbf{HiCFL} & G & \textbf{0.755}\scriptsize{$\pm$0.011} & \textbf{0.581}\scriptsize{$\pm$0.036} \\
\bottomrule
\end{tabular}
\end{center}
\caption{Results (PR-AUC) for machine learning with manual labeling.}
\label{table:ml2000}
\end{table}

\begin{table}[tp]
\begin{center}
\begin{tabular}{c|c|c|c}
\toprule
& ML-AI & ML-CS & AI-CS \\
\hline
Human & 0.698\scriptsize{$\pm$0.087} & 0.846\scriptsize{$\pm$0.074} & 0.716\scriptsize{$\pm$0.115}  \\
\hline
HiCFL & \textbf{0.854}\scriptsize{$\pm$0.017} & \textbf{0.932}\scriptsize{$\pm$0.007} & \textbf{0.768}\scriptsize{$\pm$0.023}  \\
\bottomrule
\end{tabular}
\end{center}
\caption{Accuracies of domain relevance comparison.}
\label{table:comparison}
\end{table}

\begin{table*}[ht]
\begin{center}
\resizebox{0.995\textwidth}{!}{
\begin{tabular}{|c|c|c|c|c|}
\multicolumn{5}{l}{The depth of the background color indicates the domain relevance. The darker the color, the higher the domain relevance (annotated by the authors);} \\
\multicolumn{5}{l}{* indicates the term is a core term, otherwise it is a fringe term.} \\
\hline
1-10 & 101-110 & 1001-1010 & 10001-10010 & 100001-100010 \\
\hline
\cellcolor{gray!80}{supervised learning}* & \cellcolor{gray!70}{adversarial machine learning}* & \cellcolor{gray!60}regularization strategy & \cellcolor{gray!20}method for detection & tumor region \\
\cellcolor{gray!70}{convolutional neural network}* & \cellcolor{gray!70}{temporal-difference learning}* & \cellcolor{gray!60}weakly-supervised approach & \cellcolor{gray!10}gait parameter & mutual trust \\
\cellcolor{gray!80}{machine learning}* & \cellcolor{gray!70}restricted boltzmann machine & \cellcolor{gray!50}learned embedding & \cellcolor{gray!20}stochastic method & inherent problem \\
\cellcolor{gray!80}{deep learning}* & \cellcolor{gray!70}{backpropagation through time}* & \cellcolor{gray!60}node classification problem & \cellcolor{gray!10}recommendation diversity & {healthcare system}* \\
\cellcolor{gray!80}{semi-supervised learning}* & \cellcolor{gray!70}svms & \cellcolor{gray!70}non-convex learning & numerical experiment & {two-phase}* \\
\cellcolor{gray!70}{q-learning}* & \cellcolor{gray!60}{word2vec}* & \cellcolor{gray!60}sample-efficient learning & second-order method & posetrack \\
\cellcolor{gray!80}{reinforcement learning}* & \cellcolor{gray!70}rbms & \cellcolor{gray!60}cnn-rnn model & \cellcolor{gray!20}landmark dataset & {half}* \\
\cellcolor{gray!80}{unsupervised learning}* & \cellcolor{gray!70}{hierarchical clustering}* & \cellcolor{gray!60}deep bayesian & \cellcolor{gray!20}general object detection & mfcs \\
\cellcolor{gray!70}{recurrent neural network}* & \cellcolor{gray!70}{stochastic gradient descent}* & \cellcolor{gray!60}classification score & \cellcolor{gray!20}cold-start recommendation & {borda count}* \\
\cellcolor{gray!70}{generative adversarial network}* & \cellcolor{gray!70}{svm}* & \cellcolor{gray!70}{classification algorithm}* & \cellcolor{gray!20}similarity of image & diverse way \\
\hline
\end{tabular}}
\end{center}
\caption{Ranking results for machine learning with HiCFL.}
\label{table:sample_ml}
\end{table*}

\begin{table*}[ht]
\begin{center}
\resizebox{0.995\textwidth}{!}{
\begin{tabular}{|c|c|c|c|c|}
\multicolumn{5}{l}{\textbf{\textit{Given positives}} (10): deep learning, neural network, deep neural network, deep reinforcement learning, multilayer perceptron, convolutional neural network, recurrent neural} \\
\multicolumn{5}{l}{network, long short-term memory, backpropagation, activation function.} \\
\hline
1-10 & 101-110 & 1001-1010 & 10001-10010 & 100001-100010 \\
\hline
\cellcolor{gray!80}{convolutional neural network}* & \cellcolor{gray!70}discriminative loss & \cellcolor{gray!70}multi-task deep learning & low light image & {law enforcement agency}* \\
\cellcolor{gray!80}{recurrent neural network}* & \cellcolor{gray!70}dropout regularization & \cellcolor{gray!50}self-supervision & \cellcolor{gray!10}face dataset & case of channel \\
\cellcolor{gray!80}{artificial neural network}* & \cellcolor{gray!50}{semantic segmentation}* & \cellcolor{gray!50}state-of-the-art deep learning algorithm & \cellcolor{gray!10}estimation network & {release}* \\
\cellcolor{gray!80}{feedforward neural network}* & \cellcolor{gray!70}mask-rcnn & \cellcolor{gray!50}generative probabilistic model & \cellcolor{gray!10}method on benchmark datasets & {ahonen}* \\
\cellcolor{gray!80}{deep learning}* & \cellcolor{gray!70}{probabilistic neural network}* & \cellcolor{gray!30}translation model & \cellcolor{gray!20}distributed constraint & electoral control \\
\cellcolor{gray!80}{neural network}* & \cellcolor{gray!70}pretrained network & \cellcolor{gray!30}probabilistic segmentation & \cellcolor{gray!20}gradient information & {runge}* \\
\cellcolor{gray!80}{generative adversarial network}* & \cellcolor{gray!70}discriminator model & \cellcolor{gray!40}handwritten digit classification & model on a variety & many study \\
\cellcolor{gray!80}{multilayer perceptron}* & \cellcolor{gray!60}sequence-to-sequence learning & \cellcolor{gray!60}deep learning classification & \cellcolor{gray!20}model constraint & {mean value}* \\
\cellcolor{gray!80}{long short-term memory}* & \cellcolor{gray!80}autoencoders & \cellcolor{gray!50}multi-task reinforcement learning & \cellcolor{gray!15}automatic detection & efficient beam \\
\cellcolor{gray!80}{neural architecture search}* & \cellcolor{gray!70}conditional variational autoencoder & \cellcolor{gray!60}{skip-gram}* & \cellcolor{gray!30}feature redundancy & {pvt}* \\
\hline
\end{tabular}}
\end{center}
\caption{Ranking results for deep learning with HiCFL (PU learning).}
\label{table:sample_dl}
\end{table*}

\subsection{Comparison to Baselines}

To compare with baselines, 
we separate a portion of core terms as queries for evaluation. Specifically, for each domain, we use 80\% labeled terms for training, 10\% for validation, and 10\% for testing (with automatic annotation). 
Terms in the validation and testing sets are treated as fringe terms. By doing this, the evaluation can represent the general performance for all fringe terms to some extent. 
And the model comparison is fair since the rich information of terms for evaluation is not used in training.
We also create a test set with careful human annotation on \textit{machine learning} to support our overall evaluation, which contains 2000 terms, with half for evaluation and half for testing.

As evaluation metrics, we calculate both ROC-AUC and PR-AUC with automatic or manually created labels. ROC-AUC is the area under the receiver operating characteristic curve, and PR-AUC is the area under the precision-recall curve. 
If a model achieves higher values, most of the domain-relevant terms are ranked higher, which means the model has a better measurement on the domain relevance of terms.

Table~\ref{table:results1} and Table \ref{table:results2} show the results for three broad/narrow domains respectively.
We observe our proposed CFL and HiCFL outperform all the baselines, and the standard deviations are low.
Compared to MLP, CFL achieves much better performance benefiting from the core-anchored semantic graph and feature aggregation, which demonstrates the domain relevance can be bridged via term relevance.
Compared to CFL, HiCFL works better owing to hierarchical learning.

In the PU setting-- the situation when automatic annotation is not applied to the target domain, although only 20 positives are given, HiCFL still achieves satisfactory performance and significantly outperforms all the baselines (Table~\ref{table:results3}). 

The PR-AUC scores on the manually created test set without and with the PU setting are reported in Table~\ref{table:ml2000}. We observe that the results are generally consistent with results reported in Table~\ref{table:results2} and Table~\ref{table:results3}, 
which indicates the evaluation with core terms can work just as well.

\subsection{Comparison to Human Performance}

In this section, we aim to compare our model with human professionals in measuring the fine-grained domain relevance of terms. 
Because it is difficult for humans to assign a score representing domain relevance directly,
we generate term pairs as queries and let humans judge which one in a pair is more relevant to \textit{machine learning}.
Specifically, we create 100 ML-AI, ML-CS, and AI-CS pairs respectively.
Taking ML-AI as an example, each query pair consists of an ML term and an AI term, and the judgment is considered right if the ML term is selected.

The human annotation is conducted by five senior students majoring in computer science and doing research related to terminology. 
Because there is no clear boundary between ML, AI, and CS, it is possible that a CS term is more relevant to machine learning than an AI term. 
However, the overall trend is that the higher the accuracy, the better the performance.
From Table~\ref{table:comparison}, we observe that HiCFL far outperforms human performance. 
Although we have reduced the difficulty, the task is still very challenging for human professionals.

\subsection{Case Studies}

We interpret our results by ranking terms according to their domain relevance regarding \textit{machine learning} or \textit{deep learning},
with hierarchy CS $\to$ AI $\to$ ML $\to$ DL. 
For CS-ML, we label terms with automatic annotation. 
For DL, we create 10 DL terms manually as the positives for PU learning.

Table~\ref{table:sample_ml} and Table~\ref{table:sample_dl} show the ranking results (1-10 represents terms ranked $1$st to $10$th).
We observe the performance is satisfactory. For ML, important concepts such as supervised learning, unsupervised learning, and deep learning are ranked very high. Also, terms ranked before $1010$th are all good domain-relevant terms. For DL, although only 10 positives are provided, the ranking results are quite impressive. E.g., unlabeled positive terms like artificial neural network, generative adversarial network, and neural architecture search are ranked very high. Besides, terms ranked $101$st to $110$th are all highly relevant to DL, and terms ranked $1001$st to $1010$th are related to ML.

\section{Conclusion}

We introduce and study the fine-grained domain relevance of terms-- an important property of terms that has not been carefully studied before. 
We propose a hierarchical core-fringe domain relevance learning approach, which can cover almost all terms in human languages and various domains, while requires little or even no human annotation.

We believe this work will inspire an automated solution for knowledge management and help a wide range of downstream applications in natural language processing.
It is also interesting to integrate our methods to more challenging tasks, for example, to characterize more complex properties of terms even understand terms.

\section*{Acknowledgments}

We thank the anonymous reviewers for their valuable comments and suggestions.
This material is based upon work supported by the National Science Foundation IIS 16-19302 and IIS 16-33755, Zhejiang University ZJU Research 083650, IBM-Illinois Center for Cognitive Computing Systems Research (C3SR) - a research collaboration as part of the IBM Cognitive Horizon Network, grants from eBay and Microsoft Azure, UIUC OVCR CCIL Planning Grant 434S34, UIUC CSBS Small Grant 434C8U, and UIUC New Frontiers Initiative. Any opinions, findings, and conclusions or recommendations expressed in this publication are those of the author(s) and do not necessarily reflect the views of the funding agencies.

\bibliographystyle{acl_natbib}
\bibliography{acl2021}

\end{document}